\documentclass{llncs}
\usepackage{mathtools}
\usepackage{amssymb}

\usepackage{tikz}
\usepackage{pgfplots}

\usepackage[numbers]{natbib}

\usepackage[]{algorithm2e}
\usepackage{algorithmic}

\newcommand{\vc}[1]{\mathbf{#1}}
\newcommand{\mH}{\mathcal{H}}
\newcommand{\mX}{\mathcal{X}}
\newcommand{\mY}{\mathcal{Y}}
\newcommand{\mE}{\mathbb{E}}
\newcommand{\mR}{\mathbb{R}}
\newcommand{\mN}{\mathbb{N}}
\newcommand{\bx}{\mathbf{x}}
\newcommand{\tx}{\tilde{\bx}}
\newcommand{\dd}{\mathsf{d}}
\newcommand{\by}{\mathbf{y}}
\newcommand{\ba}{\mathbf{a}}
\newcommand{\bc}{\mathbf{c}}
\newcommand{\bq}{\mathbf{q}}
\newcommand{\bz}{\mathbf{z}}
\newcommand{\bI}{\mathbf{I}}
\newcommand{\mI}{\mathcal{I}}
\newcommand{\bK}{\mathbf{K}}
\newcommand{\mS}{\mathcal{S}}
\newcommand{\bX}{\mathbf{X}}
\newcommand{\bZ}{\mathbf{Z}}
\newcommand{\pg}{\pmb{\gamma}}
\newcommand{\pe}{\pmb{\epsilon}}
\newcommand{\po}{\pmb{\omega}}
\newcommand{\pl}{\pmb{\lambda}}

\DeclareMathOperator*{\argmin}{arg\,min}%
\newcommand{\stcomp}[1]{{#1}^{\mathsf{c}}}  
\usepackage{arydshln}

\usepackage{hyperref}

\usepackage{caption}
\usepackage{subcaption}

\newif\ifShowComments
\ShowCommentstrue 

\usepackage{xcolor}

\begin{document}

\title{Large-scale Nonlinear Variable Selection via Kernel Random Features}
\titlerunning{Large-scale nonlinear variable selection}

\author{Magda Gregorov\'a\inst{1,2} \and
				Jason Ramapuram\inst{1,2} \and
        Alexandros Kalousis\inst{1,2} \and
        St\'ephane Marchand-Maillet\inst{2}
}

\institute{Geneva School of Business Administration, HES-SO, Switzerland 
\and University of Geneva, Switzerland}

\maketitle

\begin{abstract}
We propose a new method for input variable selection in nonlinear regression.
The method is embedded into a kernel regression machine that can model general nonlinear functions, not being a priori limited to additive models.
This is the first kernel-based variable selection method applicable to large datasets.
It sidesteps the typical poor scaling properties of kernel methods by mapping the inputs into a relatively low-dimensional space of random features.
The algorithm discovers the variables relevant for the regression task together with learning the prediction model through learning the appropriate nonlinear random feature maps.
We demonstrate the outstanding performance of our method on a set of large-scale synthetic and real datasets.
\end{abstract}

\section{Introduction}\label{sec:Intro}
It has been long appreciated in the machine learning community that learning sparse models can bring multiple benefits such as better interpretability, improved accuracy by reducing the curse of dimensionality, computational efficiency at prediction times, reduced costs for gathering and storing measurements, etc.
A plethora of sparse learning methods has been proposed for linear models \cite{Hastie2015}.
However, developing similar methods in the nonlinear setting proves to be a challenging task.

Generalized additive models \cite{Hastie1990} can use similar sparse techniques as their linear counterparts. However, the function class of linear combinations of nonlinear transformations is too limited to represent general nonlinear functions.
Kernel methods \cite{Scholkopf2002} have for long been the workhorse of nonlinear modelling.
Recently, a substantial effort has been invested into developing kernel methods with feature selection capabilities \cite{Bolon-Canedo2013}.
The most successful approaches within the filter methods are based on mapping distributions into the reproducing kernel Hilbert spaces (RKHS) \cite{Muandet2016}.
Amongst the embedded methods, multiple algorithms use the feature-scaling weights proposed in \cite{Weston2003}.
The authors in \cite{Rosasco2013} follow an alternative strategy based on the function and kernel partial derivatives.

All the kernel-based approaches above suffer from a common problem: they do not scale well for large data sets. 
The kernel methods allow for nonlinear modelling by applying 
high dimensional (possibly infinite-dimensional)
nonlinear transformations $\phi : \mX \to \mH$ to the input data.
Due to what is known as the kernel trick, these transformations do not need to be explicitly evaluated. 
Instead, the kernel methods operate only over the inner products between pairs of data points that can be calculated quickly by the use of positive definite kernel functions $k : \mX \times \mX \to \mR, \, k(\bx,\tx)=\langle \phi(\bx),\phi(\tx) \rangle$.
Given that these inner products need to be calculated for all data-point pairs, the kernel methods are generally costly for datasets with a large number $n$ of training points both in terms of computation and memory.
This is further exacerbated for the kernel variable selection methods, which typically need to perform the $\mathcal{O}(n^2)$ kernel evaluations multiple times (per each input dimension, or with each iterative update).

In this work we propose a novel kernel-based method for input variable selection in nonlinear regression that can scale to datasets with large numbers of training points. 
The method builds on the idea of approximating the kernel evaluations by Fourier random features \cite{Rahimi2007}.
Instead of fixing the distributions generating the random features a priori, it learns them together with the predictive model such that they degenerate for the irrelevant input dimensions.
The method falls into the category of embedded approaches that seek to improve predictive accuracy of the learned models through sparsity \cite{Kohavi1997}.
This is the first kernel-based variable selection method for general nonlinear functions that can scale to large datasets of tens of thousands of training data.

\section{Background}\label{sec:Background}

We formulate the problem of nonlinear regression as follows:
given a training set of $n$ input-output pairs $\mS_n = \{ (\bx_i,y_i) \in (\mX \times \mY) \, : \, \mX \subseteq \mR^d, \mY \subseteq \mR, \, i \in \mN_n \}$ sampled i.i.d. according to some unknown probability measure $\rho$,
our task is to estimate the regression function $f : \mX \to \mY, \, f(\bx) = \mE(y|\bx)$ that minimizes the expected squared error loss $\mathcal{L}(f) = \mathbb{E} \left(y - f(\bx) \right)^2 = \int \left(y - f(\bx) \right)^2 d\rho(\bx,y)$.

In the variable selection setting, we assume that the regression function does not depend on all the $d$ input variables.
Instead, it depends only on a subset $\mI$ of these of size $l < d$, so that $f(\bx) = f(\tx)$ if $x^s = \tilde{x}^s$ for all dimensions $s \in \mI$.

We follow the standard theory of regularised kernel learning and estimate the regression function as the solution to the following problem
\begin{equation}\label{eq:RegLearning}
\widehat{f} = \argmin_{f \in \mH} \, \mathcal{\widehat{L}}(f) + \lambda ||f||^2_\mH \enspace .
\end{equation}
Here the function hypothesis space $\mH$ is a reproducing kernel Hilbert space (RKHS), $||f||_\mH$ is the norm induced by the inner product in that space, and $\mathcal{\widehat{L}}(f) = \frac{1}{n} \sum_i^n \left( y_i - f(\bx_i) \right)^2$ is the empirical loss replacing the intractable expected loss above. 

From the standard properties of the RKHS, the classical result (e.g. \cite{Scholkopf2002}) states that the evaluation of the minimizing function $\widehat{f}$ at any point $\tx \in \mX$ can be represented as a linear combination of the kernel functions $k$ over the $n$ training points
\begin{equation}\label{eq:KernelRepresentation}
\widehat{f}(\tx) = \sum_i^n c_i \, k(\bx_i,\tx) \enspace .
\end{equation}
The parameters $\bc$ are obtained by solving the linear problem
\begin{equation}\label{eq:KernelLinearProblem}
(\bK + \lambda \bI_n) \, \bc = \by \enspace ,
\end{equation}
where $\bK$ is the $n \times n$ kernel matrix with the elements $K_{ij} = k(\bx_i,\bx_j)$ for all $\bx_i,\bx_j \in \mS_n$.

\subsection{Random Fourier Features}\label{sec:RandomFourierFeatures}

Equations \eqref{eq:KernelRepresentation} and \eqref{eq:KernelLinearProblem} point clearly to the scaling bottlenecks of the kernel regression.
In principal, at training it needs to construct and keep in memory the $(n \times n)$ kernel matrix and solve an $n$ dimensional linear system ($\propto \mathcal{O}(n^3)$).
Furthermore, the whole training set $\mS_n$ needs to be stored and accessed at test time so that the predictions are of the order $\mathcal{O}(n)$.

To address these scaling issues, the authors in \cite{Rahimi2007} proposed to map the data into a low-dimensional Euclidean space $\bz: \mX \to \mR^D$ so that the inner products in this space are close approximations of the corresponding kernel evaluation $\langle \bz(\bx), \bz(\tx) \rangle_{\mR^D} \approx \langle \phi(\bx), \phi(\tx) \rangle_{\mH} = k(\bx,\tx)$.
Using the nonlinear features $\bz(\bx) \in \mR^D$ the evaluations of the minimising function can be approximated by 
\begin{equation}\label{eq:FuncApprox}
\widehat{f}(\tx) \approx \langle \bz(\tx), \ba \rangle_{\mR^D} \enspace ,
\end{equation}
where the coefficients $\ba$ are obtained from solving the linear system
\begin{equation}\label{eq:RFLinearProblem}
(\bZ^T \bZ + \lambda \bI_D) \, \ba = \bZ^T \by \enspace ,
\end{equation}
where $\bZ$ is the $(n \times D)$ matrix of the random features for all the data points.
The above approximation requires the construction of the $\bZ$ matrix
and solving the $D$-dimensional linear problem, hence significantly reducing the training costs if $D \ll n$.
Moreover, access to training points is no longer needed at test time and the predictions are of the order $\mathcal{O}(D) \ll \mathcal{O}(n)$.

To construct well-approximating features, the authors in \cite{Rahimi2007} called upon Bochner's theorem which states that a continuous function $g : \mR^d \to \mR$ with $g(\vc{0}) = 1$ is positive definite if and only if it is a Fourier transform of some probability measure on $\mR^d$.
For translation-invariant positive definite kernels we thus have
\begin{equation}\label{eq:Bochner}
k(\bx,\tx) = g(\bx-\tx) = g(\pl) = \int_{\mR^d} e^{i \, \po^T\pl} \, \dd \mu(\po) 
\enspace ,
\end{equation}
where $\mu(\po)$ is the probability measure on $\mR^d$.
In the above, $g$ is the characteristic function of the multivariate random variable $\po$ defined by the expectation 
$g(\pl) = \mE_{\po} ( e^{i \, \po^T\pl} ) = \mE_{\po} ( e^{i \, \po^T (\bx - \tx)} ) = \mE_{\po} ( e^{i \, \po^T \bx} e^{-i \, \po^T \tx} ) = k(\bx,\tx)$.

It is straightforward to show that the expectation over the complex exponential can be decomposed into an expectation over an inner product 
$\mE_{\po} (e^{i \, \po^T (\bx - \tx)}) = \mE_{\po} \langle \psi_{\po}(\bx), \psi_{\po}(\tx) \rangle$ where the nonlinear mappings are defined as $\psi_{\po}: \mX \to \mR^2$, $\psi_{\po}(\bx) = [ \cos(\po^T\bx) \,
\sin(\po^T\bx) ]^T$.
In \cite{Rahimi2007} the authors proposed an even lower-dimensional transformation $\varphi_{\po,b}(\bx): \mX \to \mR $  
\begin{equation}\label{eq:randFeat}
\varphi_{\po,b}(\bx) = \sqrt{2} \cos(\po^T\bx + b) \enspace ,
\end{equation}
where $b$ is sampled uniformly from $[0,2\pi]$ and that satisfies the expectation equality $\mE_{\po} (e^{i \, \po^T (\bx - \tx)}) = \mE_{\po,b} \langle \varphi_{\po,b}(\bx), \varphi_{\po,b}(\tx) \rangle$.
We chose to work with the mapping $\varphi$ (dropping the subscripts $\po,b$ when there is no risk of confusion) in the remainder of the text.
The approximating nonlinear feature $\bz(\bx)$ for each data-point $\bx$ is obtained by concatenating $D$ instances of the random mappings 
$\bz(\bx) = [\varphi^1(\bx), \ldots, \varphi^D(\bx)]^T$
with $\po$ and $b$ sampled according to their probability distribution so that the expectation is approximated by the sample sum.

\subsection{Variable Selection Methods}\label{sec:VariableSelectionMethods}

In this section we position our research with respect to other nonlinear methods for variable selection
with an emphasis on kernel methods.

In the class of generalized additive models, lessons learned from the linear models can be reused to construct sparse linear combinations of the nonlinear functions of each variable or, taking into account also possible interactions, of all possible pairs, triplets, etc., e.g. \cite{Ravikumar2007,Yin2012,Lin2006,Tyagi2016}.
%
Closely related to these are the multiple kernel learning methods that seek to learn a sparse linear combination of kernel bases, e.g. \cite{Bach2008,Bach2009a,Koltchinskii2010}.
While these methods have shown some encouraging results, their simplifying additive assumption and the fast increasing complexity when higher-order interactions shall be considered (potentially $2^d$ additive terms for $d$ input variables) clearly present a serious limitation.

Recognising these shortcomings, multiple alternative approaches for general nonlinear functions were explored in the literature. 
They can broadly be grouped into three categories \cite{Kohavi1997}: filters, wrappers and embedded methods.

The filter methods consider the variable selection as a preprocessing step that is then followed by an independent algorithm for learning the predictive model.
Many traditional methods based on information-theoric or statistical measures of dependency (e.g. information gain, Fisher-score, etc.) fall into this category \cite{Bolon-Canedo2015}.
More recently, significant advancement has been achieved in formulating criteria more appropriate for non-trivial nonlinear dependencies \cite{Gretton2008,Song2007,Fukumizu2012,Yamada2014,Ren2015,Chen2017}.
These are based on the use of (conditional) cross-covariance operators arising from embedding probability measures into the reproducing kernel Hilbert spaces (RKHS) \cite{Muandet2016}.
However, they are still largely disconnected from the predictive model learning procedure and oblivious of the effects the variable selection has on the final predictive performance.

The wrapper methods perform variable selection on top of the learning algorithm treating it as a black box.
These are practical heuristics (such as greedy forward or backward elimination) for the search in the $2^d$ space of all possible subsets of the $d$ input variables \cite{Kohavi1997}.
Classical example in this category is the SVM with Recursive Feature Elimination \cite{Guyon2002}.
The wrappers are universal methods that can be used on top of any learning algorithm but they can become expensive for large dimensionalities $d$, especially if the complexity of the underlying algorithm is high.

Finally, the embedded methods link the variable selection to the training of the predictive model with the view to achieve higher predictive accuracy stemming from the learned sparsity.
Our method falls into this category.
There are essentially just two branches of kernel-based methods here:
methods based on feature rescaling \cite{Grandvalet2002,Weston2003,Rakotomamonjy2003,Maldonado2011,Allen2013}, and derivative-based methods \cite{Rosasco2013,Gregorova2018}. 
We discuss the feature rescaling methods in more detail in section \ref{sec:LinkToScaling}.
The derivative based methods use regularizers over the partial derivatives of the function and exploit the derivative reproducing property \cite{Zhou2008} to arrive at an alternative finite-dimensional representation of the function.
Though theoretically intriguing, these methods scale rather badly as in addition to the $(n \times n)$ kernel matrix they construct also the $(nd \times n)$ and $(nd \times nd)$ matrices of first and second order partial kernel derivatives and use their concatenations to formulate the sparsity constrained optimisation problem.

There exist two other large groups of \emph{sparse} nonlinear methods.
These address the sparsity in either the latent feature representation, e.g. \cite{Gurram2014}, or in the data instances, e.g. \cite{Chan2007}.
While their motivation partly overlaps with ours (control of overfitting, lower computational costs at prediction), their focus is on a different notion of sparsity that is out of the scope of our discussion.


\section{Towards Sparsity in Input Dimensions}\label{sec:TowardsSparsity}

As stated above, our objective in this paper is learning a regression function that is sparse with respect to the input variables.
Stated differently, the function shall be insensitive to the values of the inputs in the $d-l$ large complement set $\stcomp{\mI}$ of the irrelevant dimensions so that $f(\bx) = f(\tx)$ if $x^s = \tilde{x}^s$ for all $s \in \mI$.

From equation \eqref{eq:FuncApprox} we observe that the function evaluation is a linear combination of the D random features $\varphi$. 
The random features \eqref{eq:randFeat} are in turn constructed from the input $\bx$ through the inner product $\po^T \bx$.
Intuitively, if the function $\widehat{f}$ is insensitive to an input dimension $s$, the value of the corresponding input $x^s$ shall not enter the inner product $\po^T \bx$ generating the D random features.
Formally, we require $\omega^s x^s = 0$ for all $s \in \stcomp{\mI}$ which is obviously achieved by $\omega^s = 0$. 
We therefore identify the problem of learning sparse predictive models with sparsity in vectors $\po$.


\subsection{Learning Through Random Sampling}\label{sec:Learning through random sampling}

Though in equation \eqref{eq:randFeat} $\po$ appears as a parameter of the nonlinear transformation $\varphi$, it 
cannot be learned directly as it is the result of random sampling from the probability distribution $\mu(\po)$.
In order to ensure the same sparse pattern in the D random samples of $\po$, we use a procedure similar to what is known as the reparametrization trick in the context of variational auto-encoders \cite{Kingma2014}.

We begin by expanding equation \eqref{eq:Bochner} of the Bochner's theorem into the marginals across the $d$ dimensions\footnote{This is possible due to the independence of the $d$ dimensions of the r.v. $\po$.}
\begin{equation}\label{eq:Bochner2}
g(\pl) = \int_{\mR^d} e^{i \, \po^T\pl} \, \dd \mu(\po) = \int_{\mR} e^{i \, \omega^1 \lambda^1} \dd \mu(\omega^1) \ldots \int_{\mR} e^{i \, \omega^d \lambda^d} \dd \mu(\omega^d)
\enspace .
\end{equation}
To ensure that $\omega^s = 0$ when $s \in \stcomp{\mI}$ in all the D random samples, the corresponding probability measure (distribution) $\mu(\omega^s)$ needs to degenerate to $\delta(\omega^s)$.
The distribution  $\delta(\omega^s)$ has all its mass concentrated at the point $\omega^s = 0$, and has the property $\int_{\mX} h(\omega^s) \, \dd \delta(\omega^s) = h(0)$.
In particular for $h$ the complex exponential we have $\int_{\mX} e^{i \, \omega^s \lambda^s} \, \dd \delta(\omega^s) = 1$
so that the value of $\lambda^s$ has no impact on the product in equation \eqref{eq:Bochner2}, and therefore no impact on $g(\pl)$.\footnote{And from \eqref{eq:Bochner} and \eqref{eq:FuncApprox} it neither impacts the kernel and regression function evaluation.}

\subsubsection{Reparametrization trick}\label{sec:ReparametrizationTrick}
To ensure that all the D random samples of $\po$ have the same sparse pattern we need to be able to optimise through its random sampling.
%
For each element $\omega$ of the vector $\po$, we parametrize the sampling distributions $\mu_\gamma(\omega)$ by its scale $\gamma$ so that $\lim_{\gamma \to 0}$ $\mu_\gamma(\omega) = \delta(\omega)$.
We next express each of the univariate random variables $\omega$ as a deterministic transformation of the form $\omega = q_{\gamma}(\epsilon) = \gamma \epsilon$ (scaling) of an auxiliary random variable $\epsilon$ with a fixed probability distribution $\mu_1(\epsilon)$ with the scale parameter $\gamma = 1$.
For example, for the Gaussian and Laplace kernels the auxiliary distribution $\mu_1(\epsilon)$ are the standard Gaussian and Cauchy respectively.

By the above reparametrization of the random variable $\po$ we disconnect the sampling operation over $\pe$ from the rescaling operation $\po = \bq_{\pg}(\pe) = \pe \odot \pg$ with a deterministic parameter vector $\pg$.
Sparsity in $\po$ (and therefore the learned model) can now be achieved by learning sparse parameter vector $\pg$.

Though in principle it would be possible to learn the sparsity in the sampled $\po$'s directly, this would mean sparsifying instead of one vector $\pg$ the D sampled vectors $\po$.
Moreover, the procedure would need to cater for the additional constraint that all the samples have the same sparse pattern. While theoretically possible, we find our reparametrization approach more elegant and practical.

\subsection{Link to Feature Scaling}\label{sec:LinkToScaling}

In the previous section we have built our strategy for sparse learning using the inverse Fourier transform of the kernels and the degeneracy of the associated probability measures.
When we plug the rescaling operation into the random feature mapping \eqref{eq:randFeat}
\begin{equation}\label{eq:randFeat2}
\varphi(\bx) = \sqrt{2} \cos(\po^T\bx + b)  = \sqrt{2} \cos((\pe \odot \pg)^T\bx + b) = \sqrt{2} \cos(\pe^T (\pg \odot \bx) + b)
\enspace ,
\end{equation}
we see that the parameters $\pg$ can be interpreted as weights scaling the input variables $\bx$.
This makes a link to the variable selection methods based on feature scaling.
These are rather straightforward when the kernel is simply linear, or when the nonlinear transformations $\phi(\bx)$ can be evaluated explicitly (e.g. polynomial) \cite{Grandvalet2002,Weston2003}. In essence, instead of applying the weights to the input features, they are applied to the associated model parameters and suitably constrained to approximate the zero-norm problem.

More complex kernels, for which the nonlinear features $\phi(\bx)$ cannot be directly evaluated (may be infinite dimensional), are considered in \cite{Rakotomamonjy2003,Maldonado2011,Allen2013}.
Here the scaling is applied within the kernel function $k(\pg \odot \bx,\pg \odot \tx)$.
The methods typically apply a two-step iterative procedure: they fix the rescaling parameters $\pg$ and learn the corresponding $n$-long model parameters vector $\bc$ (equation \eqref{eq:KernelRepresentation}); fix $\bc$ and learn the $d$-long rescaling vector $\pg$ under some suitable zero-norm approximating constraint.
The naive formulation for $\pg$ is a nonconvex problem that requires calculating derivatives of the kernel functions with respect to $\pg$ (which depending on the kernel function may become rather expensive).
In \cite{Allen2013}, the author proposed a convex relaxation based on linearization of the kernel function.
Nevertheless, all the existing methods applying the feature scaling within the kernel functions scale badly with the number of instances as they need to recalculate the $(n \times n)$ kernel matrix and solve the corresponding optimisation (typically $\mathcal{O}(n^3)$) with every update of the weights $\pg$.

\section{Sparse Random Fourier Features Algorithm}\label{sec:Algorithm}

In this section we present our algorithm for learning with Sparse Random Fourier Features (SRFF).

\begin{algorithm}
\caption{Sparse Random Fourier Features (SRFF) algorithm}\label{alg:Alg1}
\SetKwInOut{Input}{Input}
\SetKwInOut{Output}{Output}
\SetKwInOut{Initialise}{Initialise}
\SetKwInOut{Objective}{Objective}
\SetKwBlock{Step1}{Step1}{end}
\SetKwBlock{S2}{Step2}{end}
\Input{training data $(\bX, \by)$; hyper-parameters $\lambda, D$, size of $\Delta$ simplex}
\Output{model parameters $\ba$, scale vector $\pg$}
\Initialise{$\pg$ evenly on simplex $\Delta$, $\pe_j \sim \mu_I(\pe)$ and $b_j \sim U[0,2\pi],$ $\forall j \in \mN_D$}
\Objective{$J(\ba,\pg) = ||\by-\bZ \ba||_2^2 + \lambda ||\ba||_2^2$}
\BlankLine
\Repeat(\tcp*[f]{Alternating descent}) {objective convergence}{
\Begin(Step 1: Solve for $\ba$){
rescalings $\po_j = \pg \odot \pe_j, \ \forall j \in \mN_D$ \\
random features $\bz(\bx)=[\varphi^1(\bx),\ldots,\varphi^D(\bx)], \, \forall \bx \in \mS_n$  \tcp*[f]{equation \eqref{eq:randFeat2}} \\
$\ba \leftarrow \argmin_a ||\by-\bZ \ba||_2^2 + \lambda ||\ba||_2^2$ \tcp*[f]{equation \eqref{eq:RFLinearProblem}} \\
}
\Begin(Step 2: Solve for $\pg$){
$\pg \leftarrow \argmin_{\pg \in \Delta} ||\by-\bZ \ba||_2^2$ \tcp*[f]{projected gradient descent}
}
}
\end{algorithm}

Similarly to the feature scaling methods we propose a two-step alternative procedure to learn the model parameters $\ba$ and the distribution scalings $\pg$.
For a fixed $\pg$ we generate the random features for all the input training points $\mathcal{O}(nD)$, and solve the linear problem $\eqref{eq:RFLinearProblem}$ $\mathcal{O}(D^3)$ to get the $D$-long model parameters $\ba$.
Given that in our large-sample settings we assume $D \ll n$, this step is significantly cheaper than the corresponding step for learning the $\bc$ parameters in the existing kernel feature scaling methods described in section \ref{sec:LinkToScaling}.

In the second step, we fix the model parameters $\ba$ and learn the $d$-long vector of the distribution scalings $\pg$.
We formulate the optimisation problem as the minimisation of the empirical squared error loss with $\pg$ constrained on the probability simplex $\Delta$ to encourage the sparsity.
\begin{equation}\label{eq:GammaProblem}
\argmin_{\pg \in \Delta} J(\pg), \qquad \qquad J(\pg) := || \by - \bZ \ba ||_2^2
\end{equation}
Here the $(n \times D)$ matrix $\bZ$ is constructed by concatenating the D random features $\varphi$ with the $\pg$ rescaling \eqref{eq:randFeat2}.

We solve problem \eqref{eq:GammaProblem} by the projected gradient method with accelerated FISTA line search \cite{Beck2009a}. 
The gradient can be constructed from the partial derivatives as follows
\begin{equation}\label{eq:GammaGrad}
\begin{aligned}
\frac{\partial J(\pg)}{\gamma^s} & = -(\by - \bZ \, \ba)^T \, \frac{\partial \bZ}{\partial \gamma^s} \, \ba
\qquad \qquad \forall s \in \mN_d \\
\frac{\partial Z_{ij}}{\partial \gamma^s} 
& = - \sqrt{2} \sin(\pe^T (\pg \odot \bx) + b) \, \epsilon^s x^s, \qquad \epsilon^s = \omega_s / \gamma_s \enspace .
\end{aligned}
\end{equation}
Unlike in the other kernel feature scaling methods, the form of the gradient \eqref{eq:GammaGrad} is always the same irrespective of the choice of the kernel. 
The particular kernel choice is reflected only in the probability distribution from which the auxiliary variable $\pe$ is sampled and has no impact on the gradient computations.
In our implementation (\url{https://bitbucket.org/dmmlgeneva/srff_pytorch}), we leverage the automatic differentiation functionality of pytorch in order to obtain the gradient values directly from the objective formulation.

\section{Empirical Evaluation}\label{sec:Empirical evaluation}

We implemented our algorithm in pytorch and made it executable optionally on CPUs or GPUs. 
All of our experiments were conducted on GPUs (single p100). 
The code including the settings of our experiments amenable for replication is publicly available at \url{https://bitbucket.org/dmmlgeneva/srff_pytorch}.

In our empirical evaluation we compare to multiple baseline methods.
We included the nonsparse random Fourier features method (RFF) of \citep{Rahimi2007} in our main SRFF code as a call option.
For the naive mean and ridge regression we use our own matlab implementation.
For the linear lasso we use the matlab PASPAL package \cite{Mosci2010}.
For the nonlinear Sparse Aditive Model (SPAM) \cite{Ravikumar2007} we use the R implementation of \cite{Zhao2014}.
For the Hilberth-Schmidt Independece Criterion lasso method (HSIC) \cite{Yamada2014}, and the derivative-based embedded method of \cite{Rosasco2013} (Denovas) we use the authors' matlab implementation.

Except SPAM, all of the baseline sparse learning methods use a two step procedure for arriving at the final model. They first learn the sparsity using either predictive-model-dependent criteria (lasso, Denovas) or in a completely disconnected fashion (HSIC).
In the second step (sometimes referred to as de-biasing \cite{Rosasco2013}), they use a base non-sparse learning method (ridge, or kernel ridge) to learn a model over the selected variables (including hyper-parameter search and cross-validation).
For HSIC, which is a filter method that does not natively predict the regression outputs, we use as the second step our implementation of the RFF.
It searches through the candidate sparsity patterns HSIC produces and uses the validation mean square error as a criteria for the final model selection.
In contrast to these, our SRFF method is a \emph{single step procedure} that does not necessitate this extra re-learning phase.
\subsubsection{Experimental Protocol} In all our experiments we use the same protocol.
We randomly split the data into three independent subsets: train, validation and test. 
We use the train subset for training the models, we use the validation subset to perform the hyper-parameter search, and we use the test set to evaluate the predictive performance.
We repeat all the experiments 30 times, each with a different random train/validation/test split.

We measure the predictive performance in terms of the root mean squared error (RMSE) over the test samples, averaged over the 30 random replications of the experiments. 
The regularization hyper-parameter $\lambda$ (exists in ridge, lasso, Denovas, HSIC, RFF and SRFF) is searched within a 50-long data-dependent grid (automatically established by the methods).
The smoothing parameter in Denovas is fixed to 10 following the authors' default \cite{Rosasco2013}.
We use the Gaussian kernel for all the experiments with the width $\sigma$ set as the median of the Euclidean distances amongst the 20 nearest neighbour instances.
We use the same kernel in all the kernel methods and the corresponding scale parameter $\gamma = 1/\sigma$ in the random feature methods for comparability of results.
We fix the number of random features to $D = 300$ for all the experiments in both RFF and SRFF.

We provide the results of the baseline nonlinear sparse methods (SPAM, HSIC, Denovas) only for the smallest experiments. 
As explained in the previous sections, the motivation for our paper is to address the poor scaling properties of the existing methods.
Indeed, none of the baseline kernel sparse methods scales to the problems we consider here.
HSIC \cite{Yamada2014} creates a $(n \times n)$ kernel matrix per each dimension $d$ and solves a linear lasso problem over the concatenated vectorization of these with memory requirements $(n^2 \times d)$ and complexity $\mathcal{O}(n^4)$.
In our tests, it did not finish within 24hrs running on 20 CPUs (Dual Core Intel Xeon E5-2680 v2 / 2.8GHz) for the smallest training size of 1000 instances in our SE3 experiment.
Within the same time span it did not arrive at a solution for any of the experiments with $n > 1000$.
Denovas constructs, stores in memory, and operates over the $(n \times n)$, $(nd \times n)$ and $(nd \times nd)$ kernel matrix and the matrices of the first and second order derivatives. 
In our tests the method finished with an out-of-memory error (with 32GB RAM) for the SE1 with 5k training samples and for SE2 problem already with 1k training instances.
SPAM finished with errors
for most of the real-data experiments.


\subsection{Synthetic Experiments}\label{sec:Synthetic experiments}

We begin our empirical evaluation by exploring the performance over a set of synthetic experiments.
The purpose of these is to validate our method under controlled conditions when we understand the true sparsity of the generating model.
We also experiment with various nonlinear functions and increasing data sizes in terms of both the sample numbers $n$ and the dimensionality $d$.

\renewcommand{\tabcolsep}{4pt}
\begin{table}[h!]
\begin{center}
\caption{Summary of synthetic experiments}\label{tab:Synthetic}
\begin{tabular}{l | c |  c | c | c | c }
\hline & & & & & \\ [-1.5ex] 
Exp & Train & Test & Total & Relevant & Generative \\
code & size & size & dims & dims & function \\
\hline & & & & & \\ [-1.5ex] 
SE1 & 1k - 50k & 1k & 18 & 5 & $y = \sin\left( (x_1+x_3)^2 \right)  \sin(x_7 x_8 x_9) + N(0,0.1)$\\
SE2 & 1k - 50k & 1k & 100 & 5 & $y = \log\left( (\sum_{s=11}^{15} x_s)^2 \right) + N(0,0.1)$ \\
SE3 & 1k - 50k & 10k & 1000 & 10 & $y = 10(z_1^2 + z_3^2) e^{-2(z_1^2 + z_3^2)} + N(0,0.01)$\\
\hline
\end{tabular}
\end{center}
{\footnotesize We use the same size for the test and validation samples. 
In all the experiments, the data instances are generated from a standard normal distribution.
In the functions, subscripts are dimensions, superscripts are exponents.
For more detailed description of the generative function of SE3 see the appropriate section in the text.}
\end{table}

\subsubsection{SE1:}\label{sec:SE1}
The very first of our experiments is a rather small problem with only $d=18$ input dimensions of which only 5 are relevant for the regression function.
In Table \ref{tab:SE1sample1k} we compare our SRFF method to the baselines for the smallest sample setting with $n = 1000$.
Most of the methods (linear, additive or non-sparse) do not succeed in learning a model for the complex nonlinear relationships between the inputs and outputs and fall back to predicting simple mean.

\begin{figure}[h!]
\centering
\begin{tikzpicture}[scale=1]
    \begin{axis}[
            ybar,
            title style={at={(0.4,1.2)},
                anchor=north},
            title={SE1 - Learned sparsity by SRFF},
            major grid style={draw=white},
            axis x line*=bottom,
    		    axis y line*=left,
		        y axis line style={opacity=0},
            bar width=.15cm,
            width=\textwidth,
            height=3cm,
            legend style={at={(0.8,1)},
                anchor=north,legend columns=-1},
            symbolic x coords={d1,d2,d3,d4,d5,d6,d7,d8,d9,d10,d11,d12,d13,d14,d15,d16,d17,d18},
            nodes near coords align={vertical},
            ymin=0,
        ]
        \addplot table[x=interval,y=carS]{\mydata};
        \addplot table[x=interval,y=carL]{\mydata};
        \legend{1k,50k}
    \end{axis}
\end{tikzpicture}
\caption{Learned sparsity pattern $\pg$ by the SRFF method for the 1k and 50k training size in the SE1 experiment (the median of the 30 replications). The other nonlinear sparse methods learn the same pattern for the 1k problem but cannot solve the 50k problem.}\label{fig:SE1_sparsity}
\end{figure}
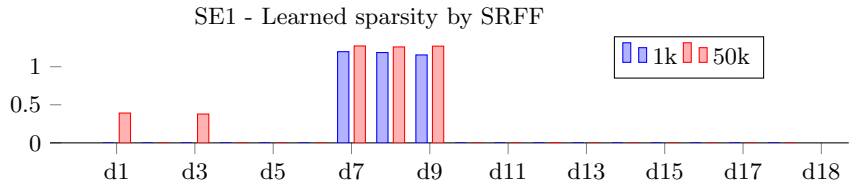

The general nonlinear models with sparsity (HSIC, Denovas and SRFF) divert from the simple mean prediction.
They all discover and use in the predictive model the same sparse pattern (see Fig. \ref{fig:SE1_sparsity} for SRFF).
Denovas and SRFF achieve almost identical results which confirms that our method is competitive with the state of the art methods in terms of predictive accuracy and variable selection.\footnote{The low predictive performance of HSIC is the result of the 2nd model fitting step. It could potentially be improved with an additional kernel learning step. However, as we keep the kernel fixed for all the other methods, we do not perform the kernel search for HSIC either.}

\begin{table}[h!]
\begin{center}
\caption{SE1 - Test RMSE for $n = 1000$}\label{tab:SE1sample1k}
\begin{tabular}{l| c | c |  c | c | c | c | c | c }
\hline & & & & & & & & \\ [-1.5ex] 
& Mean & Ridge & Lasso & RFF & SPAM & HSIC & Denovas & SRFF \\
\hline & & & & & & & & \\ [-1.5ex] 
RMSE & 0.287 & 0.287 & 0.287 & 0.287 & 0.0287 & 0.341 & \textbf{0.272} & \textbf{0.272} \\
std & 0.009 & 0.009 & 0.009 & 0.009 & 0.009 & 0.060 & 0.009 & 0.009 \\
\hline
\end{tabular}
\end{center}
{\footnotesize Predictive performance in terms of root mean squared error (RMSE)  over independent test sets for the SE1 dataset with training size $n = 1000$. The std line is the standard deviation of the RMSE across the 30 resamples.}
\end{table}

In Table \ref{tab:SE1increasing} we document how the increasing training size contributes to improving the predictive performance even in the case of several thousands instances. 
The performance of the SRFF  method for the largest 50k sample is by about $6\%$ better than for the 1k training size.
For the other methods the problem remains out of their reach\footnote{The class of linear functions is too limited and the nonlinear function with all the variables considered by RFF is too complex.} and they stick to the mean prediction even for higher training sizes.\footnote{The small variations in the error stem from using different training sets to estimate the mean.}
We do not provide any comparisons with the nonlinear sparse methods because, as explained above, they do not scale to the sample sizes we consider here.

\begin{table}[h!]
\begin{center}
\caption{SE1 - Test RMSE for increasing train size $n$}\label{tab:SE1increasing}
\begin{tabular}{ c | c | c | c | c |  c }
\hline & & & & & \\ [-1.5ex] 
$n$ & Mean & Ridge & Lasso & RFF & SRFF \\
\hline & & & & & \\ [-1.5ex] 
1k & 0.287 \scriptsize{(0.009)} & 0.287 \scriptsize{(0.009)} & 0.287 \scriptsize{(0.009)} & 0.287 \scriptsize{(0.009)} & \textbf{0.272} \scriptsize{(0.009)} \\
5k & 0.284 \scriptsize{(0.011)} & 0.284 \scriptsize{(0.011)} & 0.284 \scriptsize{(0.011)} & 0.284 \scriptsize{(0.011)} & \textbf{0.263} \scriptsize{(0.010)} \\
10k & 0.285 \scriptsize{(0.010)} & 0.285 \scriptsize{(0.010)} & 0.285 \scriptsize{(0.010)} & 0.286 \scriptsize{(0.010)} & \textbf{0.261} \scriptsize{(0.011)} \\
50k & 0.283 \scriptsize{(0.010)} & 0.283 \scriptsize{(0.010)} & 0.283 \scriptsize{(0.010)} & 0.283 \scriptsize{(0.010)} & \textbf{0.255} \scriptsize{(0.009)} \\
\hline
\end{tabular}
\end{center}
{\footnotesize Predictive performance in terms of root mean squared error (RMSE)  over independent test sets for the SE1 dataset with increasing training size $n$. The standard deviation of the RMSE across the 30 resamples is in the brackets.}
\end{table}

The improved predictive performance for the larger training sizes goes hand in hand with the variable selection, Figure \ref{fig:SE1_sparsity}.
For the smallest 1k training sample, SRFF identifies only the 7th, 8th and 9th relevant dimensions. 
They enter the sine in the generative function in a product and therefore have a larger combined effect on the function outcome than the squared sum of dimensions 1 and 3.
These two dimensions are picked up by the method from the larger training sets and this contributes to the increase in the predictive performance.

\pgfplotstableread[row sep=\\,col sep=&]{
interval & carS & carL \\
d1 & 0 & 0.39159 \\
d2 & 0 & 0 \\
d3 & 0 & 0.3786 \\
d4 & 0 & 0 \\
d5 & 0 & 0 \\
d6 & 0 & 0 \\
d7 & 1.1975 & 1.2727 \\
d8 & 1.1862 & 1.2598 \\
d9 & 1.1547 & 1.2697 \\
d10 & 0 & 0 \\
d11 & 0 & 0 \\
d12 & 0 & 5.9681e-05 \\
d13 & 0 & 0 \\
d14 & 0 & 0 \\
d15 & 0 & 0 \\
d16 & 0 & 0 \\
d17 & 0 & 0 \\
d18 & 0 & 0 \\
    }\mydata

\subsubsection{SE2:}\label{sec:SE2}
In the second experiment we increase the dimensionality to $d=100$ and change the nonlinear function (see Table \ref{tab:Synthetic}).
The overall outcomes are rather similar to the SE1 experiment.
Again, it's only the nonlinear sparse models that predict something else than mean, SPAM marginally better, HSIC marginally worse.
Our SRFF method clearly outperforms all the other methods in the predictive accuracy.
It also correctly discovers the 5 relevant variables with the median value of $\gamma$ for these dimensions between $0.92-1.04$ while the maximum for all the irrelevant variables is $0.06$.\footnote{SPAM and HSIC discover the correct patterns as well but it does not help their predictive accuracy.}
The advantage of SRFF over the baselines for large sample sizes (Table \ref{tab:SE2increasing}) is even more striking than in the SE1 experiment.

\begin{table}[h!]
\begin{center}
\caption{SE2 - Test RMSE for $n = 1000$}\label{tab:SE2sample1k}
\begin{tabular}{l | c | c |  c | c | c | c | c }
\hline & & & & & & & \\ [-1.5ex] 
& Mean & Ridge & Lasso & RFF & SPAM & HSIC & SRFF \\
\hline &  & & & & & & \\ [-1.5ex] 
RMSE & 2.216 & 2.216 & 2.216 & 2.216 & 2.162 & 2.357 & \textbf{1.603} \\
std & 0.105 & 0.105 & 0.105 & 0.104 & 0.110 & 0.141 & 0.104 \\
\hline
\end{tabular}
\end{center}
{\footnotesize Predictive performance in terms of root mean squared error (RMSE)  over independent test sets for the SE2 dataset with training size $n = 1000$. The std line is the standard deviation of the RMSE across the 30 resamples.}
\end{table}

\begin{table}[h!]
\begin{center}
\caption{SE2 - Test RMSE for increasing train size $n$}\label{tab:SE2increasing}
\begin{tabular}{ c | c | c | c | c |  c }
\hline & & & & & \\ [-1.5ex] 
$n$ & Mean & Ridge & Lasso & RFF & SRFF \\
\hline & & & & & \\ [-1.5ex] 
1k & 2.216 \scriptsize{(0.105)} & 2.216 \scriptsize{(0.105)} & 2.216 \scriptsize{(0.105)} & 2.216 \scriptsize{(0.105)} & \textbf{1.603} \scriptsize{(0.104)} \\
5k & 2.211 \scriptsize{(0.079)} & 2.211 \scriptsize{(0.079)} & 2.211 \scriptsize{(0.079)} & 2.211 \scriptsize{(0.079)} & \textbf{1.278} \scriptsize{(0.076)} \\
10k & 2.224 \scriptsize{(0.115)} & 2.224 \scriptsize{(0.115)} & 2.224 \scriptsize{(0.115)} & 2.224 \scriptsize{(0.115)} & \textbf{1.272} \scriptsize{(0.138)} \\
50k & 2.224 \scriptsize{(0.082)} & 2.224 \scriptsize{(0.082)} & 2.224 \scriptsize{(0.082)} & 2.224 \scriptsize{(0.082)} & \textbf{1.273} \scriptsize{(0.080)} \\
\hline
\end{tabular}
\end{center}
{\footnotesize Predictive performance in terms of root mean squared error (RMSE)  over independent test sets for the SE2 dataset with increasing training size $n$. The standard deviation of the RMSE across the 30 resamples is in the brackets.}
\end{table}
\subsubsection{SE3:}\label{sec:SE3}
In this final synthetic experiment we increase the dimensionality to $d=1000$ to further stretch our SRFF method.
There are only 10 relevant input variables in this problem.
The first 5 were generated as random perturbations of the random variable $z_1$, e.g. $x_1 = z_1 + N(0,0.1)$, the second 5 by the same procedure from $z_2$, e.g. $x_5 = z_2 + N(0,0.1)$.
The remaining 990 input variables were generated by the same process from the other 198 standard normal $z$'s.

We summarise the results for the 1k and 50k training instances in Table \ref{tab:SE3increasing}.
As in the other synthetic experiments, the baseline methods are not able to capture the nonlinear relationships of this extremely sparse problem and instead predict a simple mean.
Our SRFF method achieves significantly better accuracy for the 1k training set, and it further considerably improves with 50k samples to train on.
These predictive gains are possible due to SRFF correctly discovering the set of relevant variables. 
In the 1k case, the medians across the 30 data resamples of the learned $\pg$ parameters are between $0.37-0.71$ for the 10 relevant variables and maximally $0.05$ for the remaining 990 irrelevant variables.
In the 50k case, the differences are even more clearly demarcated: $1.19-1.64$ for the relevant, and $0.03$ maximum for the irrelevant (bearing in mind that the total sum over the vector $\pg$ is the same in both cases).

\begin{table}[h!]
\begin{center}
\caption{SE3 - Test RMSE for increasing train size $n$}\label{tab:SE3increasing}
\begin{tabular}{ c | c | c | c | c |  c }
\hline & & & & & \\ [-1.5ex] 
$n$ & Mean & Ridge & Lasso & RFF & SRFF \\
\hline & & & & & \\ [-1.5ex] 
1k & 0.676 \scriptsize{(0.002)} & 0.676 \scriptsize{(0.002)} & 0.676 \scriptsize{(0.002)} & 0.676 \scriptsize{(0.002)} & \textbf{0.478} \scriptsize{(0.031)} \\
50k & 0.677 \scriptsize{(0.002)} & 0.677 \scriptsize{(0.002)} & 0.677 \scriptsize{(0.002)} & 0.677 \scriptsize{(0.002)} & \textbf{0.206} \scriptsize{(0.004)} \\
\hline
\end{tabular}
\end{center}
{\footnotesize Predictive performance in terms of root mean squared error (RMSE)  over independent test sets for the SE3 dataset with increasing training size $n$. The standard deviation of the RMSE across the 30 resamples is in the brackets.}
\end{table}

\vspace{-2em}
\subsection{Real Data Experiments}\label{sec:RealData}

\begin{table}[h!]
\begin{center}
\caption{Summary of real-data experiments}\label{tab:RealData}
\begin{tabular}{l | c | c |  c | c | c }
\hline & & & & & \\ [-1.5ex] 
Data & Dataset & Exp & Train & Test & Total\\
source & name & code & size & size & dims\\
\hline & & & & & \\ [-1.5ex] 
LIAC & Computer Activity & RCP & 6k & 1k & 21 \\
LIAC & F16 elevators & REL & 6k & 1k & 17 \\
LIAC & F16 ailernos & RAI & 11k & 1k & 39 \\
Kaggle & Ore mining impurity & RMN & 50k & 10k & 21 \\
\hline
\end{tabular}
\end{center}
{\footnotesize We use the same size for the test and validation samples.}
\end{table}

We experiment on four real datasets: three from the LIACC\footnote{\url{http://www.dcc.fc.up.pt/~ltorgo/Regression/DataSets.html}} regression repository, and one Kaggle dataset\footnote{\url{https://www.kaggle.com/edumagalhaes/quality-prediction-in-a-mining-process}}.
The summary of these is presented in Table \ref{tab:RealResults}.
The RFF results illustrate the advantage nonlinear modelling has over simple linear models. 
Our sparse nonlinear SRFF method clearly outperforms all the linear as well as the non-sparse nonlinear RFF method. Moreover, it is the only nonlinear sparse learning method that can handle problems of these large-scale datasets.

\begin{table}[h!]
\begin{center}
\caption{Real-data experiments - Test RMSE for increasing train size $n$}\label{tab:RealResults}
\begin{tabular}{ c | c | c | c | c |  c }
\hline & & & & & \\ [-1.5ex] 
$n$ & Mean & Ridge & Lasso & RFF & SRFF \\
\hline & & & & & \\ [-1.5ex] 
RCP & 18.518 \scriptsize{(0.988)} & 9.686 \scriptsize{(0.705)} & 9.689 \scriptsize{(0.711)} & 8.194 \scriptsize{(0.635)} & \textbf{2.516} \scriptsize{(0.184)} \\
REL & 1.044 \scriptsize{(0.050)} & 0.514 \scriptsize{(0.210)} & 0.468 \scriptsize{(0.178)} & 0.446 \scriptsize{(0.036)} & \textbf{0.314} \scriptsize{(0.032)} \\
RAI & 1.013 \scriptsize{(0.034)} & 0.430 \scriptsize{(0.018)} & 0.430 \scriptsize{(0.017)} & 0.498 \scriptsize{(0.038)} & \textbf{0.407} \scriptsize{(0.022)} \\
RMN & 1.014 \scriptsize{(0.006)} & 0.987 \scriptsize{(0.006)} & 0.987 \scriptsize{(0.006)} & 0.856 \scriptsize{(0.009)} & \textbf{0.716} \scriptsize{(0.008)} \\
\hline
\end{tabular}
\end{center}
{\footnotesize Predictive performance in terms of root mean squared error (RMSE)  over independent test sets for the real datasets. The standard deviation of the RMSE across the 30 resamples is in the brackets.}
\end{table}

\section{Summary and Conclusions}\label{sec:Summary}

We present here a new kernel-based method for learning nonlinear regression function with relevant variable subset selection.
The method is unique amongst the state of the art as it can scale to tens of thousands training instances, way beyond what any of the existing kernel-based methods can handle.
For example, while none of the tested sparse method worked over datasets with more than 1k instances, the CPU version of our SRFF finished the \emph{full} validation search over 50 hyper-parameters $\lambda$ in the 50k SE1 experiment within two hours on a laptop with a Dual Intel Core i3 (2nd Gen) 2350M / 2.3 GHz and 16GB RAM.

We focus here on nonlinear regression but the extension to classification problems is straightforward by replacing appropriately the objective loss function.
We used the Gaussian kernel for our experiments as one of the most popular kernels in practice.
But the principals hold for other shift-invariant kernels as well, and the method and the algorithm can be applied to them directly as soon as the corresponding probability measure $\mu(\po)$ is recovered and the reparametrization of $\po$ explained in section \ref{sec:Learning through random sampling} can be applied.

\subsubsection{Acknowledgements} This work was partially supported by the research projects HSTS (ISNET) and RAWFIE \#645220 (H2020).
The computations were performed at University of Geneva on the Baobab and Whales clusters. We specifically wish to thank Yann Sagon, the Baobab administrator, for his excellent work and continuous support.


%
%

\newpage

\bibliographystyle{splncs03}
\bibliography{srff_ECML2018_clean}

\begin{thebibliography}{10}
\providecommand{\url}[1]{\texttt{#1}}
\providecommand{\urlprefix}{URL }

\bibitem{Allen2013}
Allen, G.I.: {Automatic Feature Selection via Weighted Kernels and
  Regularization}. Journal of Computational and Graphical Statistics  (2013)

\bibitem{Bach2008}
Bach, F.: {Consistency of the group lasso and multiple kernel learning}.
  Journal of Machine Learning Research  (2008)

\bibitem{Bach2009a}
Bach, F.: {High-Dimensional Non-Linear Variable Selection through Hierarchical
  Kernel Learning}. ArXiv 0909.0844  (2009)

\bibitem{Beck2009a}
Beck, A., Teboulle, M.: {A Fast Iterative Shrinkage-Thresholding Algorithm for
  Linear Inverse Problems}. SIAM Journal on Imaging Sciences  (2009)

\bibitem{Bolon-Canedo2013}
Bol{\'{o}}n-Canedo, V., S{\'{a}}nchez-Maro{\~{n}}o, N., Alonso-Betanzos, A.: {A
  review of feature selection methods on synthetic data}. Knowledge and
  Information Systems  (2013)

\bibitem{Bolon-Canedo2015}
Bol{\'{o}}n-Canedo, V., S{\'{a}}nchez-Maro{\~{n}}o, N., Alonso-Betanzos, A.:
  {Recent advances and emerging challenges of feature selection in the context
  of big data}. Knowledge-Based Systems  (2015)

\bibitem{Chan2007}
Chan, A.B., Vasconcelos, N., Lanckriet, G.R.G.: {Direct convex relaxations of
  sparse SVM}. In: International Conference on Machine Learning (2007) (2007)

\bibitem{Chen2017}
Chen, J., Stern, M., Wainwright, M.J., Jordan, M.I.: {Kernel Feature Selection
  via Conditional Covariance Minimization}. Advances in Neural Information
  Processing Systems (NIPS)  (2017)

\bibitem{Fukumizu2012}
Fukumizu, K., Leng, C.: {Gradient-based kernel method for feature extraction
  and variable selection}. In: Advances in Neural Information Processing
  Systems (NIPS) (2012)

\bibitem{Grandvalet2002}
Grandvalet, Y., Canu, S.: {Adaptive scaling for feature selection in SVMs}. In:
  Advances in Neural Information Processing Systems (NIPS) (2002)

\bibitem{Gregorova2018}
Gregorov{\'{a}}, M., Kalousis, A., Marchand-Maillet, S.: {Structured nonlinear
  variable selection}. In: Conference on Uncertainty in Artificial Intelligence
  (UAI) (2018)

\bibitem{Gretton2008}
Gretton, A., Fukumizu, K., Teo, C.H., Song, L., Sch{\"{o}}lkopf, B., Smola,
  A.J.: {A kernel statistical test of independence}. In: Advances in Neural
  Information Processing Systems (NIPS) (2008)

\bibitem{Gurram2014}
Gurram, P., Kwon, H.: {Optimal sparse kernel learning in the empirical kernel
  feature space for hyperspectral classification}. IEEE Journal of Selected
  Topics in Applied Earth Observations and Remote Sensing  (2014)

\bibitem{Guyon2002}
Guyon, I., Weston, J., Barnhill, S., Vapnik, V.: {Gene Selection for Cancer
  Classification using Support Vector Machines}. Machine Learning  (2002)

\bibitem{Hastie1990}
Hastie, T., Tibshirani, R.: {Generalized additive models}. Chapman and Hall
  (1990)

\bibitem{Hastie2015}
Hastie, T., Tibshirani, R., Wainwright, M.: {Statistical Learning with
  Sparsity: The Lasso and Generalizations}. CRC Press (2015)

\bibitem{Kingma2014}
Kingma, D.P., Welling, M.: {Auto-Encoding Variational Bayes}. In: International
  Conference on Learning Representations (ICLR) (2014)

\bibitem{Kohavi1997}
Kohavi, R., John, G.H.: {Wrappers for feature subset selection}. Artificial
  Intelligence  (1997)

\bibitem{Koltchinskii2010}
Koltchinskii, V., Yuan, M.: {Sparsity in multiple kernel learning}. Annals of
  Statistics  (2010)

\bibitem{Lin2006}
Lin, Y., Zhang, H.H.: {Component selection and smoothing in multivariate
  nonparametric regression}. Annals of Statistics  (2006)

\bibitem{Maldonado2011}
Maldonado, S., Weber, R., Basak, J.: {Simultaneous feature selection and
  classification using kernel-penalized support vector machines}. Information
  Sciences  (2011)

\bibitem{Mosci2010}
Mosci, S., Rosasco, L., Santoro, M., Verri, A., Villa, S.: {Solving structured
  sparsity regularization with proximal methods}. In: European Conference on
  Machine Learning and Principles and Practice of Knowledge Discovery in
  Databases (ECML/PKDD) (2010)

\bibitem{Muandet2016}
Muandet, K., Fukumizu, K., Sriperumbudur, B., Sch{\"{o}}lkopf, B.: {Kernel Mean
  Embedding of Distributions: A Review and Beyond}. Foundations and Trends in
  Machine Learning  (2017)

\bibitem{Rahimi2007}
Rahimi, A., Recht, B.: {Random features for large-scale kernel machines}. In:
  Advances in Neural Information Processing Systems (NIPS) (2007)

\bibitem{Rakotomamonjy2003}
Rakotomamonjy, A.: {Variable Selection Using SVM-based Criteria}. Journal
  ofMachine Learning Research  (2003)

\bibitem{Ravikumar2007}
Ravikumar, P., Liu, H., Lafferty, J., Wasserman, L.: {Spam: Sparse additive
  models}. In: Advances in Neural Information Processing Systems (NIPS) (2007)

\bibitem{Ren2015}
Ren, S., Huang, S., Onofrey, J.A., Papademetris, X., Qian, X.: {A Scalable
  Algorithm for Structured Kernel Feature Selection.} Aistats  (2015)

\bibitem{Rosasco2013}
Rosasco, L., Villa, S., Mosci, S.: {Nonparametric sparsity and regularization}.
  Journal of Machine Learning Research  (2013)

\bibitem{Scholkopf2002}
Sch{\"{o}}lkopf, B., Smola, A.J.: {Learning with kernels}. The MIT Press (2002)

\bibitem{Song2007}
Song, L., Smola, A., Gretton, A., Borgwardt, K.M., Bedo, J.: {Supervised
  feature selection via dependence estimation}. Proceedings of the 24th
  international conference on Machine learning - ICML '07  (2007)

\bibitem{Tyagi2016}
Tyagi, H., Krause, A., Eth, Z.: {Efficient Sampling for Learning Sparse
  Additive Models in High Dimensions}. International Conference on Artificial
  Intelligence and Statistics (AISTATS)  (2016)

\bibitem{Weston2003}
Weston, J., Elisseeff, A., Scholkopf, B., Tipping, M.: {Use of the Zero-Norm
  with Linear Models and Kernel Methods}. Journal of Machine Learning Research
  (2003)

\bibitem{Yamada2014}
Yamada, M., Jitkrittum, W., Sigal, L., Xing, E.P., Sugiyama, M.:
  {High-dimensional feature selection by feature-wise kernelized Lasso.} Neural
  Computation  (2014)

\bibitem{Yin2012}
Yin, J., Chen, X., Xing, E.P.: {Group Sparse Additive Models}. In:
  International Conference on Machine Learning (ICML) (2012)

\bibitem{Zhao2014}
Zhao, T., Li, X., Liu, H., Roeder, K.: {CRAN - Package SAM}  (2014)

\bibitem{Zhou2008}
Zhou, D.X.: {Derivative reproducing properties for kernel methods in learning
  theory}. Journal of Computational and Applied Mathematics  (2008)

\end{thebibliography}

\end{document}